# Memory and Recall of Color Images Using Neural Networks-1

## - The Cue-Ball Model based on the Associative Memory Mechanism -

**Hiroshi Inazawa**
*Center for Education in Information Systems, Kobe Shoin Women's University*[1],
*1-2-1 Shinohara-Obanoyama, Nada Kobe 657-0015, Japan.*
E-mail: inazawa706@gmail.com

**Abstract**

This paper presents a neural network model for storing and recalling color images of different sizes using the Cue Ball-Recall Net systems [1–5]. The Cue Ball system consists of a Recall Net composed of multiple memory neurons each responsible for memorizing a single, distinct memory and a large number of neurons that represent image patterns. The images used in this study [6] consist of ten images (see Appendix). The pixel values of these images were converted into digital values[2] and used for subsequent processing. Accordingly, the Cue Ball contains ten Cue neurons, corresponding to the ten images stored in memory. The number of pixels varies among the images, and therefore, the number of Recall neurons comprising the Recall Net also varies accordingly. In this study, the number of Recall neurons was prepared between 106,134 and 360,000, depending on the number of pixels in each image. In this simulation, after the connection weights ($w_{ji}^{p}$、 $v_{ij}^{p}$) had been learned, we have used the learning and recall algorithms to investigate whether the Cue neuron which had memorized a specific image could be identified by the Recall Net when that a memorized image has been presented. Therefore, the most important simulation result is which Cue neuron responds when a memorized image is presented to the Recall Net. If the Cue neuron that memorizes the presented image responds, its output can be used to obtain the corresponding image data. By adding a BMP header to the resulting image data, the original memorized image can be accurately recalled.

## 1. Introduction

This paper presents a model for memorizing and recalling color images of different sizes. The model used in this study is the Cue Ball–Recall Net [1–5]. The system consists of the Cue Ball, a group of neurons in which each neuron is responsible for memorizing one memory, and the Recall Net, which comprises a large number of neurons used to represent image patterns. This study builds on a series of papers published from 2018 to 2026 [1–5]. Now, we focus on the storage and recall of color images with different image sizes. Now, I will briefly revisit the basic structure of the model. It consists of a neural network, comprising numerous neurons arranged in a two-dimensional layout to represent images, along with a set of neurons designed to memorize individual images, with each image represented by a single neuron. The neural network used to represent images is referred to as the Recall Net, while the group of neurons responsible for storing memories is called the Cue Ball. The neurons constituting the Recall Net are called Recall neurons, and those constituting the Cue Ball are called Cue neurons. Ten images of different sizes [6] are used in this study (see Appendix). Accordingly, ten Cue neurons are used to store the images.

---

[1] I was affiliated with Kobe Shoin Women's University but retired upon reaching the mandatory retirement age at the end of March 2025. Please note that as of April 2025, the university has been renamed "Kobe Shoin University."

[2] Regarding the program used in the simulation, I consulted the explanation for ChatGPT and Gemini 3, a form of generative AI.

Until now, the term "memory" in the context of neural networks has referred to associative memory models [3-8,17]. Since this model also pertains to memory, it can be classified as an associative memory model. Among associative memory models, the Hopfield-type model is particularly well-known. Other types, such as auto-associative memory models and hetero-associative memory models, also exist. In autocorrelation and cross-correlation models, the "memory rate" plays a crucial role in enabling the storage of various types of information [8-14]. This depends on the number of neurons being used, and the key lies in maximizing the ratio of the number of stored items to the total number of neurons. The model presented here falls, in principle, into the category of cross-correlation-type associative memory models. In the Cue Ball-Recall Net model described in this paper (as well as in previous papers), a single Cue neuron, which serves as the basis for memory, memorizes one individual image. Consequently, 10 images are memorized across 10 Cue neurons. In this sense, the memory rate is 100%. However, when the Recall neurons that make up the Recall net are taken into account, the memory rate decreases. Therefore, if the number of Cue neurons (the number of images to be memorized) becomes overwhelmingly large, the memory rate—defined as (number of memorized images / total number of neurons used) × 100—will, in principle, approach 100%. On the other hand, as the number of images to be memorized increases and as the size of the individual images grows, the amount of memory required also increases. Despite this drawback, given the trend toward ever-greater memory capacities in modern devices, the approach likely remains sufficiently practical. In the following, Section 2 detail the model, Section 3 present the simulation methodology and results, and Section 4, the final section, offer conclusions and discussion.

## 2. Specification of model

Although an overview of the model has already been provided in the Introduction, a more detailed description is presented this section, with the simulations in Section 3 in mind.

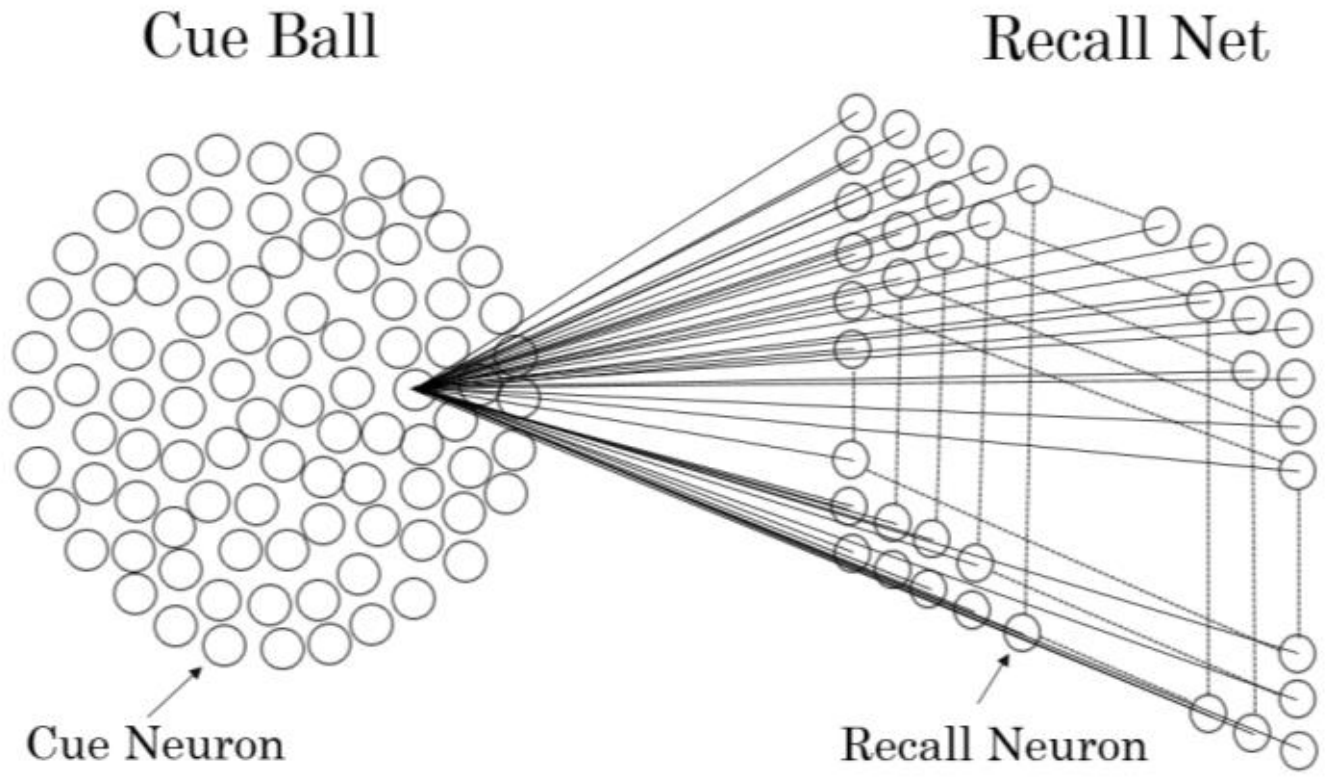

Figure.1: Diagram of the proposed model's structure. A single Cue neuron in the Cue Ball is connected to the Recall neurons in all of the Recall Nets. Using this structure, the pixel values of a color image presented to the Recall Net are memorized in digital form within a single Cue neuron.

In Figure 1, each Cue neuron is connected to all the Recall neurons in the recall network. Since 10 color images are used, 10 Cue neurons are required; meanwhile, because the 10 images vary in size, ranging from 106,134 to 360,000 pixels, up to 360,000 Recall neurons are needed. On the other hand, since Cue neurons can be added at any time, one could theoretically include an infinite number of them; however, as this study focuses on fundamental principles, the number has been set to 10, matching the number of color images used. Now, we briefly describe the connectivity of the neurons in the Cue Ball and the Recall Net. The Cue neurons in the Cue Ball are not interconnected. Similarly, the Recall neurons in the Recall Net are not interconnected. On the other hand, the Cue neuron and the corresponding Recall neuron have bidirectional connections. Note that the general Gradient Descent Method (GDM) is used as a learning method for memorization [14-16].

At this point, let us address the process of learning (encoding). The learning sequence begins with the learning of the connection weights $w_{ji}^{p}$ for the Recall neurons, followed by the learning of the connection weights $v_{ij}^{p}$ for the Cue neurons.

Let us begin by explaining the learning of the connection weights $w_{ji}^{p}$ in the Recall Net. The input-output relationships for the Cue neurons and Recall neurons are defined below.

$$y_j^p = w_{ji}^p x_i^p \ , \qquad j = 0 \sim (R-1) \tag{1}$$

where, $p$ denotes the image index, $y_j^p$ is the output of the $j$-th Recall neuron, $x_i^p$ is the input to the $i$-th Cue neuron, and $w_{ji}^p$ is the connection weight from the $i$-th Cue neuron to the $j$-th Recall neuron. In Equation (1), no summation is performed over the index "$i$". Therefore, for each image, there is only one $x_i^p$ term. In the expression "$j = 0 \sim (R-1)$", $R$ represents the number of neurons within the Recall Net. During the learning process of the Recall neurons, the digital values of one digitally encoded image are first input to the individual Recall neurons in the Recall Net (i.e., the image is presented to the Recall Net). Using these digital values and $y_j^p$ from Equation (1), the weights $w_{ji}^p(t)$ are learned via GDM. Here, the error function $E^p$ used during the learning process is defined as follows:

$$E^p \equiv \frac{1}{2}\sum_{j=0}^{R-1}\left(d_j^p - y_j^p\right)^2 \quad , \tag{2}$$

where, $d_j^p$ represents the $j$-th digital element value of the presented image $p$ input into the Recall Net. Using Equation (2), the connection weight $w_{ji}^p$ is updated via GDM as follows:

$$\begin{aligned} w_{ji}^p(t+1) &= w_{ji}^p(t) + \varDelta w_{ji}^p(t)\,, \\ \varDelta w_{ji}^p(t) &\equiv -\varepsilon_W \frac{\partial E^p}{\partial w_{ji}^p(t)} = \varepsilon_W \left(d_j^p - y_j^p(t)\right) x_i^p(t) \quad , \end{aligned} \tag{3}$$

where, $t$ is the number of updates, and $\varepsilon_W$ is the learning rate. The learning of $w_{ji}^p$ is completed using Eq. (3). Setting $\varepsilon_W$, Equations. (1) and (3) can be used to rewrite $w_{ji}^p$ and $y_j^p$ as follows.

$$\begin{aligned} w_{ji}^p(t+1) &= w_{ji}^p(t) + \left(d_j^p - w_{ji}^p(t)x_i^p(t)\right)x_i^p(t) \\ &= w_{ji}^p(t)\left(1.0 - (x_i^p(t))^2\right) + d_j^p x_i^p(t) \\ &= d_j^p\,, \end{aligned} \tag{4}$$

where Equation (1), the following equation is obtained.

$$y_j^p(t+1) = w_{ji}^p(t+1)x_i^p(t+1) = d_j^p \tag{5}$$

It follows that, after learning the connection weights $w_{ji}^p$, the output value $y_j^p$ of the Recall neuron matches the corresponding element value $d_j^p$ of the presented pattern.

Next, we describe the learning process for the connection weights $v_{ij}^p$ on the Cue neuron side. The input–output relation between the Recall neurons in the Recall Net, where one previously learned image is presented, and the $i$-th Cue neuron is defined as follows. Now, $q_i^p$ denotes the intermediate output of the Cue neuron.

$$q_i^p = \sum_{j=0}^{R-1} v_{ij}^p y_j^p, \qquad i = p = 0 \sim (N_P - 1), \tag{6}$$

where, $v_{ij}^p$ represents the connection weight from the $j$-th Recall neuron to the $i$-th Cue neuron. $N_P (= 10)$ is the number of patterns to be memorized. The final output $x_i^p$ of the Cue neuron is defined as follows, using the "$max$" function:

$$x_i^p = max(q_i^p) = \begin{cases} 1.0 & \text{for } maximum\ of\ q_i^p \\ 0.0 & \text{for } others \end{cases} \quad , \tag{7}$$

The $max$ function identifies the maximum value among the intermediate outputs $q_i^p$ of all cue neurons within a cue ball. The final output $x_i^p$ of the cue neuron with the highest intermediate output is set to 1.0, while the final outputs of all other cue neurons are set to 0.0.

The error function used in the learning process of $v_{ij}^p$ is defined as follows.

$$e^p \equiv \frac{1}{2}\sum_{i=0}^{N_p-1}\left(\theta - q_i^p\right)^2 \quad , \tag{8}$$

where, $\theta$ denotes the teacher signal for the Cue neurons and is common to all Cue neurons. Learning is considered complete when $\theta$ becomes equal to $q_i^p$.

Next, we consider the process of updating the connection weights $v_{ij}^p$ using GDM. The weights are updated according to the following equation.

$$\begin{aligned} v_{ij}^p(t'+1) &= v_{ij}^p(t') + \varDelta v_{ij}^p(t')\,, \\ \varDelta v_{ij}^p(t') &= -\varepsilon_V \frac{\partial e^p}{\partial v_{ij}^p} = \varepsilon_V \left(\theta - q_i^p(t')\right) y_j^p(t')\,, \end{aligned} \tag{9}$$

where, $t'$ denotes the update iteration, and $\varepsilon_V$ is the learning rate. The input–output Equation (6) and the learning Equation (9) demonstrate that the Cue neuron under consideration can learn the image.

First, by setting $\varepsilon_V = 1.0$, the term $q_i^p$ during the learning process (where $y_j^p$ in Equation (6) is equal to $d_j^p$) can be expressed as follows:

$$\begin{aligned} q_i^p(t'+1) &= \sum_j^{R-1} v_{ij}^p(t') + \left(\theta - q_i^p(t') d_i^p(t')\right) d_j^p(t') \\ &= \sum_j^{R-1} v_{il}^p(t') d_l^p(t') \left(1 - \sum_j^{R-1} (d_i^p(t'))^2\right) + \theta \sum_j^{R-1} (d_i^p(t'))^2 \quad , \end{aligned} \tag{10}$$

where, we set the following normalization condition for the digital value $d_j^p$ of the input image.

$$normalize_{d_j^p} = d_j^p \Big/ \sum_{k=0}^{R-1} (d_k^p)^2 \quad , \tag{11}$$

Using Equation (11) yields the following result.

$$\sum_{j=0}^{R-1} (normalized_d_j^p)^2 = 1 \tag{12}$$

In the following, this "$normalize_d_j^p$" is treated as "$d_j^p$".

$$\sum_{k=0}^{R-1} (d_k^p)^2 \equiv 1 \tag{13}$$

Using Equation (13), $q_i^p$ in Equation (10) can be expressed as follows.

$$q_i^p(t'+1) = \theta \quad . \tag{13}$$

In this way, $q_i^p(t'+1)$ becomes equal to the teacher signal $\theta$, and the learning process is completed.

## 3. Simulation and Results

This section describes the simulations based on the model presented in Section 2. In this model, only ten neurons are required for memory storage, corresponding to the number of images, whereas 106,134 to 360,000 neurons are required to represent the images, depending on their sizes. Figure.2 shows a sample of an image and the corresponding digital values for a portion of that image. The 10 color images are provided in the Appendix [6].

| Picture | Digital Data(Partial) |
|---|---|
| 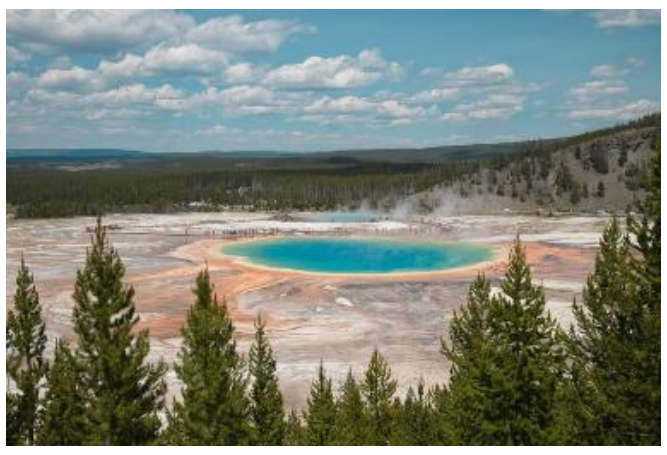 | width =399<br>height =266<br>bits =24<br>File Number=0 : 24bits Color data<br>1118982 922117 724741 1250828 1316358 ···<br>1579785 1448457 1316618 1777684 1316616 ···<br>1316357 1579785 1447945 2171669 1842702 ···<br>········· |

Figure.2: Images used in the simulation process and a portion of their digital data. The left side shows the color image used (0.bmp), while the right side displays the digital data corresponding to the initial part of that image.

The image shown in Figuer.2 is 399 pixels wide and 266 pixels high, giving a total of 106,134 pixels. Since each pixel is represented by one digital value, 106,134 Recall neurons are required to represent the image data. Note that this size is the smallest image among those used in this study. In the simulations, the connection weights $w_{ji}^p$ and $v_{ij}^p$ are trained sequentially for each image. That is, for the $p$-th image, the weights $w_{ji}^p$ and $v_{ij}^p$ are trained consecutively. However, for clarity, the two learning processes are described separately below. The learning algorithms for memorizing (encoding) $w_{ji}^p$ and $v_{ij}^p$ are summarized below.

[The learning algorithm for $w_{ji}^p$]

1. Select the images sequentially, starting from the 0-th image.
2. Read the data containing the digital values $d_j^p$ converted from the image pixel values.
3. Normalize the digital values $d_j^p$ read in Step 2 using Equation (11).
4. Initialize the connection weights of the Recall neurons: $w_{ji}^p = 1.0$.
5. Initialize $x_i^p$ and $y_j^p$, and calculate the output using Equation (1): $x_i^p = 1.0, y_j^p = 0.0$.
6. Learn the connection weights $w_{ji}^p$ according to Equation (3), using a learning rate of $\varepsilon_W = 1.0$.
7. Write the post-learning $w_{ji}^p$ to a file.

8. Return to Step 1 and repeat Steps 2–8 for each image (1–9).
9. Terminate the process after all image patterns have been learned.

[The learning algorithm for $v_{ij}^p$]

1. Select the images sequentially, starting from the 0-th image.
2. Initialize the connection weights of the Cue neurons: $v_{ij}^p$.
3. Initialize $q_j^p$ and $x_i^p$: $q_j^p = 0.0,\ x_i^p = 1.0$.
4. Load the learned connection weights $w_{ji}^p$.
5. Calculate $y_j^p$ using Equation (1) and the learned connection weights $w_{ji}^p$.
6. Using the values of $y_j^p$ obtained from Step 5, calculate $q_j^p$ in Equation (6).
7. Using the values of $y_j^p$ obtained in Step 5 and $q_j^p$ obtained in Step 6, perform the learning of $v_{ij}^p$ in Equation (9):

   *The teacher signal is set to ($\theta = 100$). This value has been determined empirically by examining the simulation results.
8. Write the post-learning $v_{ij}^p$ to a file.
9. Return to Step 1 and repeat Steps 2–9 for each image (1–9).
10. Terminate the process after all image patterns have been learned.

After the learning process, we examine whether the images have been stored in the Cue neurons. In the recall process, the digital data of an arbitrary memorized image are presented to the Recall Net, and the output values of the Recall neurons are fed into all Cue neurons in the Cue Ball. In this process, the Cue neurons that hold the presented image are expected to maximize the value of $q_i^p$. Next, we examine this recall process. The algorithm for the recall process is summarized below.

[Recall Algorithm]

1. Load the learned connection weights $w_{ji}^p$ and $v_{ij}^p$.
2. Select a desired image from a set of images that have already been processed.
3. Input the digital data $d_j^p$ of the image selected in Step 2 into the Recall neurons and calculate $q_i^p$ according to Equation (6), where $y_j^p \equiv d_j^p$. The normalized values of $d_j^p$ are used.
4. Set $q_j^p = max(q_j^p) \Rightarrow x_i^p = 1.0$, and $q_j^p < max(q_j^p) \Rightarrow x_i^p = 0.0$.
5. Select the Cue neuron for which $x_i^p = 1.0$ among those indexed from $i = 0$ to $9$.
6. For the selected Cue neuron, calculate $w_{ji}^p$ using Equation (1) and the learned connection weights $w_{ji}^p$, and reconstruct the image data.

   *At this stage, the data used for normalization is employed to restore $y_j^p$ to its original, unnormalized values.
7. Perform steps 2 through 6 above while changing the selected image.

8. Once Step 7 is complete, the process ends.

According to the learning and recall algorithms described above, once the connection weights have been trained, one of the memorized images is presented to the Recall Net. If the Cue neuron that stores the presented image can be identified, Equation (1) can be applied to that Cue neuron to accurately recall the original image. In the simulations, the Cue neuron number ($i$) correspond to the image number ($p$). However, the Cue Ball itself does not know knowledge of which image will be presented. Therefore, when a memorized image has been presented to the Recall Net, the most important factor is which Cue neuron responds. The recall process described above was applied to all memorized image data, and the output values $q_i^p$ of each Cue neuron were examined for each presented image. The results are shown in Figure.3.

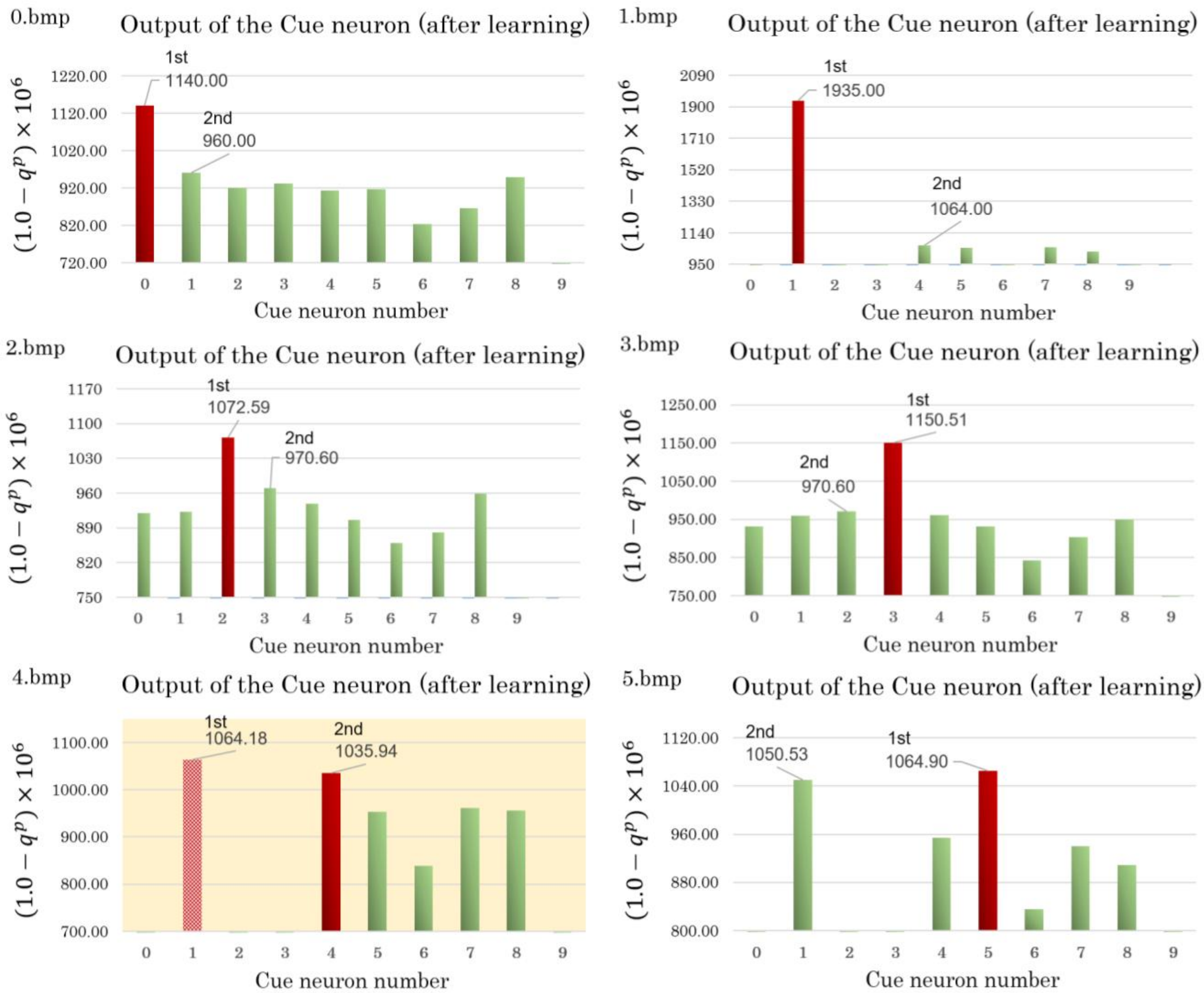

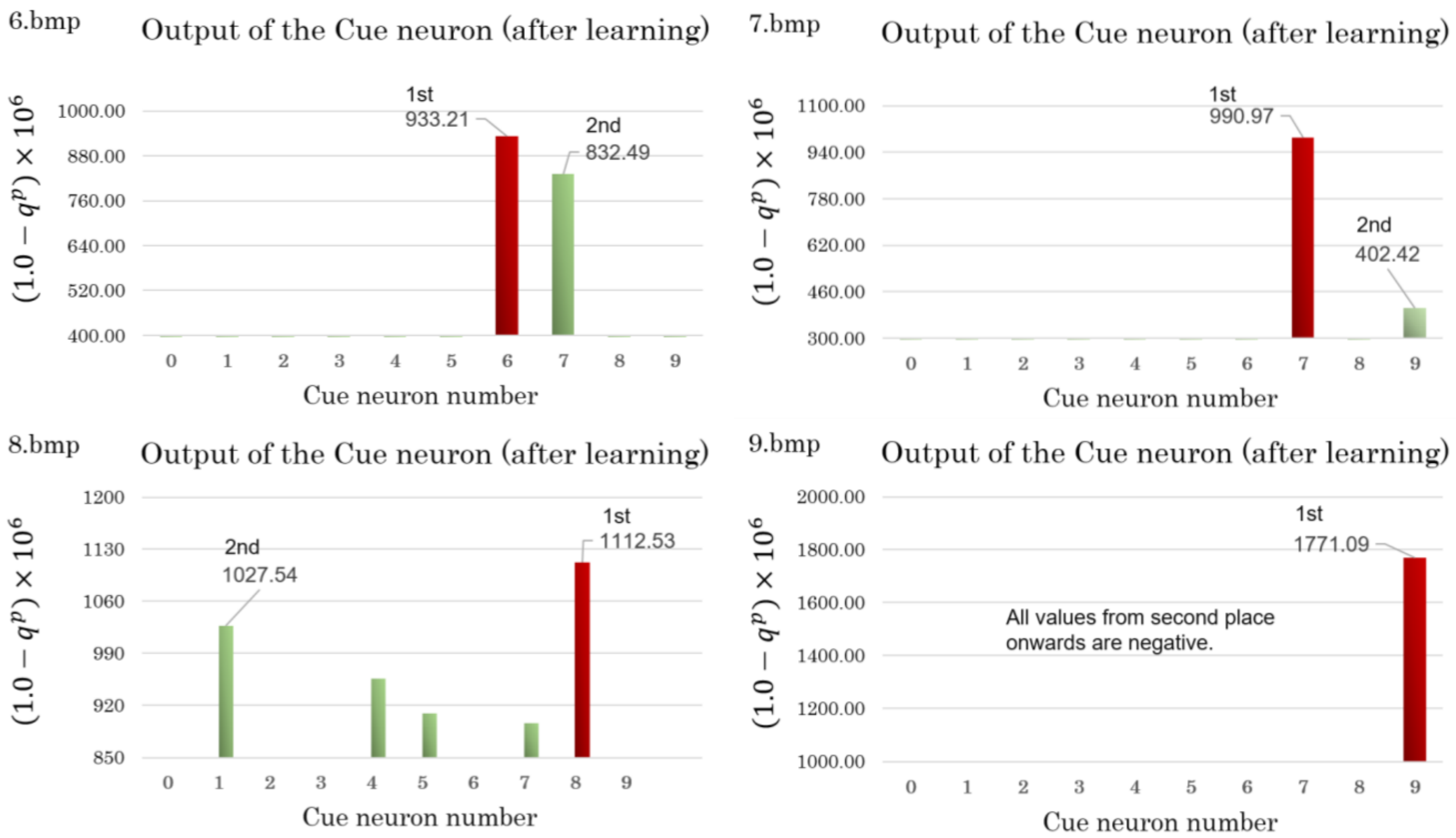

Figure.3: Output values $q_j^p$ of the Cue neurons when a stored image is presented to the Recall Net, after learning $w_{ji}^p$ and $v_{ij}^p$. The vertical axis represents the output value $(1.0 - q_j^p) \times 10^6$, and the horizontal axis represents the cue neuron number. Note that the cue neuron number and image number match. The reddish-brown filled bar graph is the cue neuron that is learning the presented image. The light green bars are other cue neurons. For example, in the case of the image "0.bmp," the output of the 0th cue neuron is the highest at 1072.59 (ranking 1st), which indicates that the memory has been successfully recalled. Meanwhile, the output value of the ranking 2nd, 970.60, corresponds to the 1st cue neuron. In contrast, the image "4.bmp" is incorrectly recalled, as indicated by the light-yellow background in the graph. In this case, the image "4.bmp" should be recalled, but the system instead recalls the image "1.bmp." The difference between the $q_j^p$ values for the incorrectly recalled and correctly recalled images was 28.24. All other images were recalled correctly.

In Figure.3, the vertical axis represents the output value $(1.0 - q_j^p) \times 10^6$, and the horizontal axis represents the Cue neuron index. Note that the Cue neuron indices correspond to the image indices. The reddish-brown filled bars represent the Cue neurons that have learned the presented (learning) images, while the light-green bars represent the other Cue neurons. For example, suppose that the image "0.bmp" is memorized in the 0th cue neuron and that its recall is successful. This Cue neuron exhibits a maximum output value of 1072.59, which ranks first. The second-largest output value, 970.60, is produced by the 1st cue neuron, which corresponds to the image "1.bmp." In contrast, the image "4.bmp" appears to have been incorrectly recalled as "1.bmp." In this instance, 4th Cue neuron, which corresponds to image "4.bmp", shows an output value of 1035.94 (the 2nd-highest value), while the 1st Cue neuron shows an output value of 1064.18 (the highest value). The latter is indicated by the reddish-brown shaded bar in the graph. The difference between the output values for the incorrectly recalled and correctly recalled images was 28.24. Although the difference was small, the output value for "1.bmp" was slightly higher than that for "4.bmp." A visual inspection of the images in the Appendix suggests that, although the exact reason is unclear, the similar portions of the sky in the two images may have contributed to this misrecall. As for the other images, however, recall has been successful in all cases.

Using the image numbers for the Cue neurons obtained here, the images were reconstructed by appending a BMP header to the image data derived from Equation (1). All of the reconstructed images were identical to the original images. It is of interest to investigate how the learned Cue neurons respond when an incomplete version of a memorized image, in which a portion of the image is missing, is presented to the Recall Net. Since this investigation requires extensive analysis, we plan to report the results in a subsequent paper.

## 4. Conclusions and Discussion

This paper presented a model for storing and recalling color images of different sizes. The underlying neural network model, referred to as the Cue Ball–Recall Net system, is based on a series of papers published from 2018 to 2026 [1–5]. In this study, ten Cue neurons were used in the Cue Ball to memory the images, corresponding to the number of images to be memorized. The Recall Net was composed of a number of Recall neurons equal to the number of pixels in each image, ranging from 106,134 to 360,000. The ten images [6] used in this study are presented in the Appendix. In the simulations, the pixel values of these color images were converted into digital values for processing.

In the simulations, after learning the connection weights $w_{ji}^p$ and $v_{ij}^p$ using the proposed learning and recall algorithms, we have investigated whether the Cue neuron responsible for storing a particular image could be identified when that image has been presented to the Recall Net. Therefore, when an image is presented to the Recall Net, the crucial information for the simulation results lies in which Cue neuron produces the maximum output value ($q_j^p$). If the Cue neuron that memories the presented image responds, the original stored image can be accurately recalled by applying Equation (1) to the output of that Cue neuron. Figure.3 shows the results of an investigation into the responses of Cue neurons in this scenario. For the stored images "0.bmp" through "9.bmp," recall was successful for all images except "4.bmp." In other words, the cue neuron corresponding to each memorized image exhibited the highest output value. It appears that the system mistakenly have retrieved "1.bmp" instead of "4.bmp." The difference in the output values $q_j^p$ between the incorrectly have retrieved image and the correctly retrieved image has been 28.24. Although this difference is small, the output value for "1.bmp" has been slightly higher than that for "4.bmp." The $q_j^p$ value associated with this misrecall depends on the synaptic connection weights $v_{ij}^p$ of the Cue neurons. However, since numerous connection weights are involved, the details of this relationship remain unclear. It should also be noted that, using the identified Cue neuron indices, we have reconstructed the images by adding BMP headers to the output values $y_i^p$ obtained from Equation (1), together with the values used for normalization in Equation (12). All reconstructed images have been identical to the original images.

As mentioned in Section 3, the numerical values of $q_j^p$ shown in Figure. 3 depend on the teacher signal $\theta$ in Equation (9).
So far, no analytical method has been found for determining the teacher signal $\theta$. Therefore, the only available approach is an empirical one, in which various values of $\theta$ are tested. Some of the results obtained using this approach are presented in Table.1.

| Cue Neuron No. | 0.bmp | | |
|---|---|---|---|
| | θ=100 | θ=10000 | θ=10 |
| | $q_i^p$ | | |
| 0 | 1.001140 | 1.115157 | 1.000104 |
| 1 | 1.000960 | 1.096915 | 1.000087 |
| 2 | 1.000920 | 1.092887 | 1.000084 |
| 3 | 1.000931 | 1.094043 | 1.000085 |
| 4 | 1.000913 | 1.092240 | 1.000083 |
| 5 | 1.000917 | 1.092591 | 1.000083 |
| 6 | 1.000823 | 1.083129 | 1.000075 |
| 7 | 1.000865 | 1.087398 | 1.000079 |
| 8 | 1.000949 | 1.095867 | 1.000086 |
| 9 | 1.000404 | 1.040797 | 1.000037 |

Table.1. Changes in the $q_j^p$ values of the learned Cue neurons for three teacher signal values: $\theta = 100, \theta = 10{,}000$, and $\theta = 10$. Now, the changes in the $q_j^p$ values for "0.bmp" are shown. Although the values of $q_j^p$ (the fractional components) fluctuate significantly, the cue neuron with the highest rank remains invariant for all values of $\theta$ (the 1st-ranked $q_j^p$ is the Cue neuron index=0). Now, $\theta = 100$ was used in Figure. 3.

Table.1 shows the variation in the $q_j^p$ values for "0.bmp" across three different teacher signal values ($\theta = 100, \theta = 10000$, and $\theta = 10$). Although the decimal portions of the $q_j^p$ values fluctuate significantly, the cue neuron ranked 1st remains the same for all images, regardless of the teacher signal value. In contrast, variations are observed in the rankings from 2nd place onwards. Within the range of values investigated, no definitive value of $\theta$ was found to be optimal. It is of interest to investigate how the learned Cue neurons respond when an incomplete version of a memorized image, with a portion of the image missing, is presented to the Recall Net. One advantage of memorizing images and other patterns in a neural network as an associative memory model is its ability to recall the original image even when an incomplete version is presented. Research on memory recall using incomplete images requires vast amounts of simulation data and yields a large volume of results; therefore, these findings will be presented in a future report.

A drawback of this model is that the number of neurons increases as the number of stored images or their size grows, which in turn leads to higher memory consumption. In this study, the number of images has been also kept relatively small. As a result, the memory required to store the connection weights $w_{ji}^p$ and $v_{ij}^p$ has been limited to approximately 24.93 MB each, for a total of 49.86 MB.

## Acknowledgments

The simulations and manuscript preparation for this study were carried out in a general study booth at the "Kobe University Library for Human Sciences." The author would like to express sincere gratitude for providing a place to conduct this research.

## Appendix

The ten photographs used in this study are shown below. The labels beneath the images, such as "0.bmp," indicate the corresponding image file names.

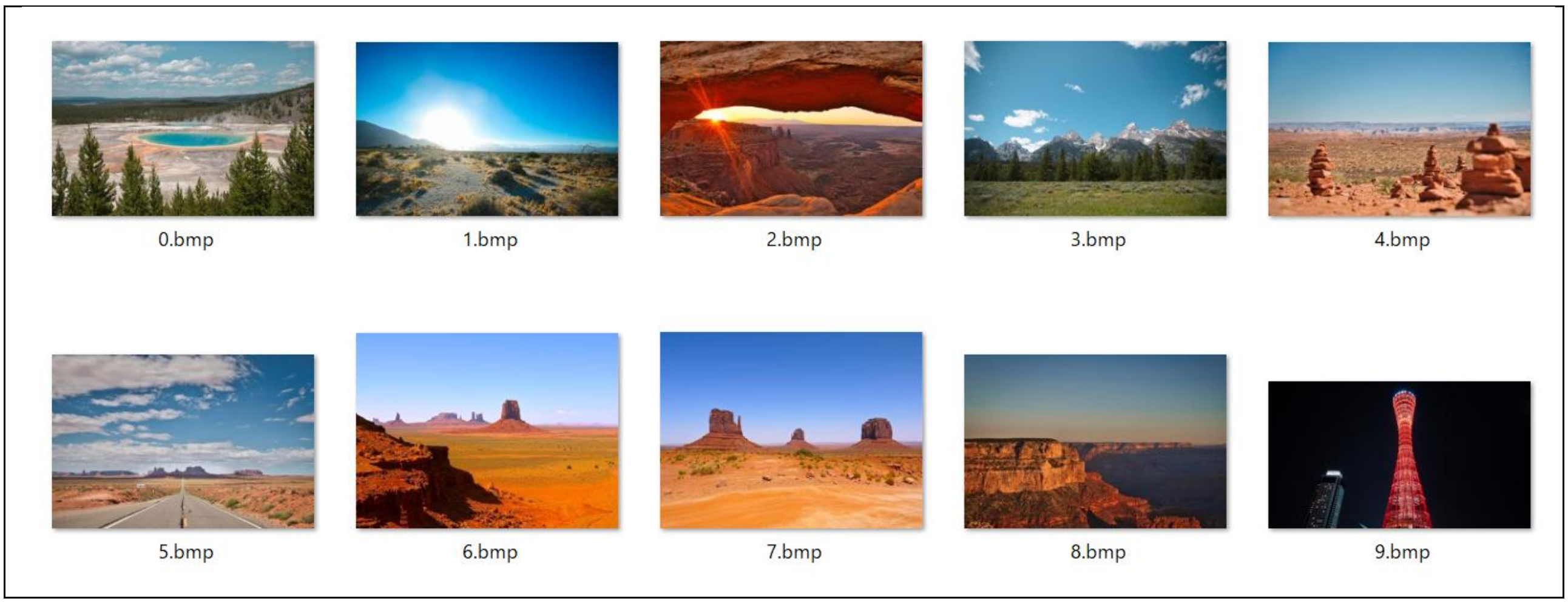